\documentclass[10pt,twocolumn,letterpaper]{article}

\usepackage{cvpr}
\usepackage{tikz}
\usetikzlibrary{shapes,arrows,calc}
\usepackage{times}
\usepackage{epsfig}
\usepackage{graphicx}
\usepackage{amsmath}
\usepackage{amssymb}
\usepackage{pdfpages}

\usepackage{times}
\usepackage{epsfig}
\usepackage{graphicx}
\usepackage{nccmath}

\usetikzlibrary{shapes.geometric}
\usetikzlibrary{positioning}
\usetikzlibrary{fadings}
\usetikzlibrary{calc}

\usepackage{color}
\usepackage{pgfplots}
\pgfplotsset{compat=1.14}
\usepackage{pgf-umlsd}
\usepackage{ifthen}
\usepackage{calc}

\usepackage[font=small,labelfont=bf,tableposition=top]{caption}

\usepackage{cuted}
\usepackage{capt-of}
\usepackage{array}

\usepackage[normalem]{ulem}
\usepackage[export]{adjustbox}

\usepackage[breaklinks=true,bookmarks=false]{hyperref}

\cvprfinalcopy 

\def\cvprPaperID{968} 

\begin{document}

\title{Pose-Guided High-Resolution Appearance Transfer via Progressive Training}

\author{Ji Liu \footnotemark\\
Carnegie Mellon University\\
{\tt\small jiliu@andrew.cmu.edu}
\and
Heshan Liu \footnotemark[\value{footnote}]\\
Carnegie Mellon University\\
{\tt\small heshanl@andrew.cmu.edu}
\and
Mang-Tik Chiu\\
UIUC\\
{\tt\small mtchiu2@illinois.edu}
\and
Yu-Wing Tai\\
Tencent \\
{\tt\small yuwingtai@tencent.com}
\and
Chi-Keung Tang\\
HKUST\\
{\tt\small cktang@cs.ust.hk}
}
\maketitle
\footnotetext{$^\ast$ Equal contribution. Authorship order was determined by rolling dice. This research was supported by the Research Grant Council, Hong Kong SAR under grant no. 16201420.}
\begin{abstract}
We propose a novel pose-guided appearance transfer network for transferring a given reference appearance to a target pose in unprecedented image resolution ($1024^2$), given respectively an image of the reference and target person. No 3D model is used. Instead, our network utilizes dense local descriptors including local perceptual loss and local discriminators to refine details, which is trained progressively in a coarse-to-fine manner to produce the high-resolution output to faithfully preserve complex appearance of garment textures and geometry, while hallucinating seamlessly the transferred appearances including those with dis-occlusion. Our progressive encoder-decoder architecture can learn the reference appearance inherent in the input image at multiple scales. Extensive experimental results on the Human3.6M dataset, 
the DeepFashion dataset, 
and our dataset collected from YouTube show that our model produces high-quality images, which can be further utilized in useful applications such as garment transfer between people and pose-guided human video generation.
\end{abstract}
\begin{figure}
    \input{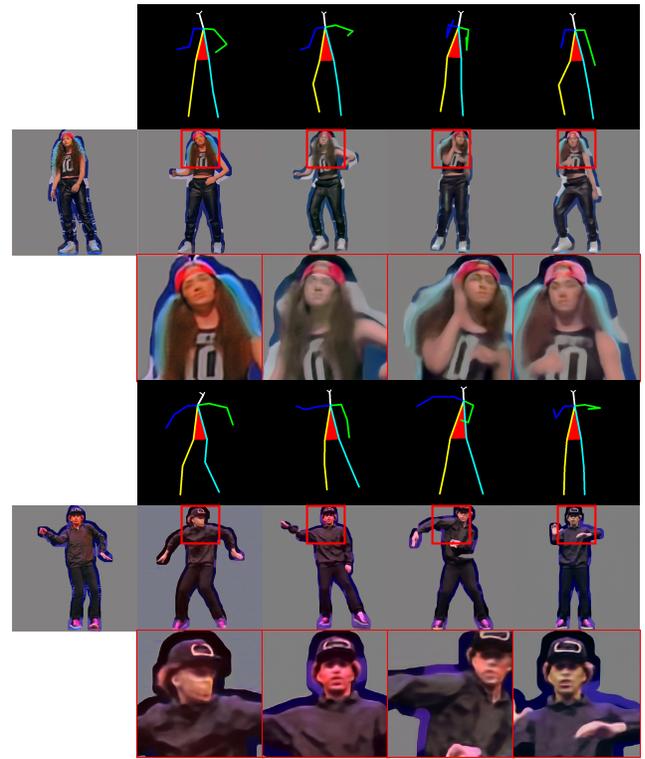}
\vspace{-0.2in}
    \caption{\textbf{Pose transfer on YouTube dataset.} Test results on our self-collected high-resolution ($1024^2$) dataset. Given a reference image (leftmost column) and target poses as input which contains self-occlusion with complex appearance in texture and geometry, our method transfers the reference appearance to target pose in high resolution while faithfully preserving complex appearance and facial features under large pose variations.}
    \label{fig:youtube}
\vspace{-0.2in}
\end{figure}
\begin{figure}[t]
    \input{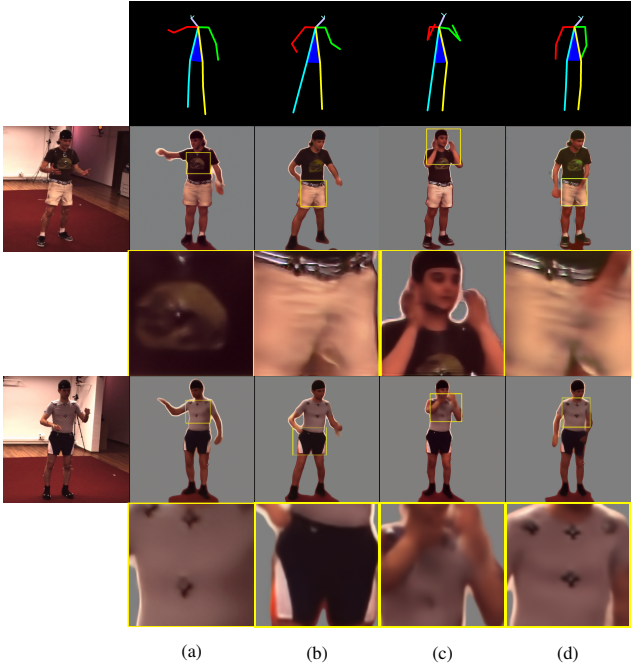}
\vspace{-0.2in}
    \caption{\textbf{Pose transfer on Human3.6M dataset.}  Test results on human3.6M dataset. (a), (b) demonstrate the effect of dis-occlusion; (c), (d) demonstrate the effect of transferring to other self-occluding poses with zoom-in views. }
    \label{fig:teaser}
\vspace{-0.2in}
\end{figure}

\section{Introduction}
Learning 3D appearance from 2D images is challenging because
images are 2D projections of the corresponding 3D world where objects can undergo complex
deformation and occlusion.
Existing relevant work in computer vision
and machine learning either uses some forms of 3D model to produce quality results,
or produces relatively low-resolution output image if only 2D images are allowed.
\par
This paper focuses on images of human, whose different poses introduce complex
non-rigid deformation and self-occlusion. 
Specifically, given a reference image of a person, our method seamlessly transfers the reference appearance to the person at the target pose while preserving high-quality garment texture of the reference person, and at the same time hallucinating realistically their complex appearance under the target pose, see Figures~\ref{fig:youtube} and~\ref{fig:teaser}.  Note that the
network should not only move the corresponding body parts to match the target pose,
but also realistically inpaint or hallucinate exposed body/garment parts unseen
in the input due to occlusion.
This is particularly challenging for human images due to the non-rigid nature
of 3D human body and complex texture and geometry distortion on 3D garment
worn by humans.\par
To address these challenges, we propose to learn appearance information
inherent in a given reference image and transfer the original appearance
of the person according to the target pose representation, by injecting the target pose representation into
the bottleneck of the encoder-decoder architecture. More importantly, to transfer detailed appearance, we enforce both global and local loss to encourage the network to learn both global coherency and local details. In addition, progressive growth is employed on both encoder and decoder to increase output resolution.
To validate our approach, we conduct extensive experiments on the Human3.6M~\cite{Ionescu2014Human36MLS} dataset, the DeepFashion~\cite{Liu2016DeepFashionPR} dataset and our dataset collected from YouTube. We
apply our method to other applications such as high-quality garment transfer and pose-guided
human video generation, demonstrating its huge potential in many
challenging tasks.\par
Our contribution consists of a new encoder-decoder architecture
that successfully enables appearance transfer to a target pose.
To enable high-resolution appearance transfer, 1) we propose novel
local descriptors (progressive local perceptual loss + local discriminators at the highest
resolution ($1024^2$)  to enhance local details and generation quality;
2) we apply progressive training to our autoencoder architecture to achieve outputs
at unprecedented high resolution ($1024^2$). To our knowledge, this is the
first progressive, deep encoder-decoder transfer network that can realistically
hallucinate in such high resolution at the target pose the complex appearance of
the worn garment, including the portion that was previously occluded in the
reference image.

\begin{figure*}
    \input{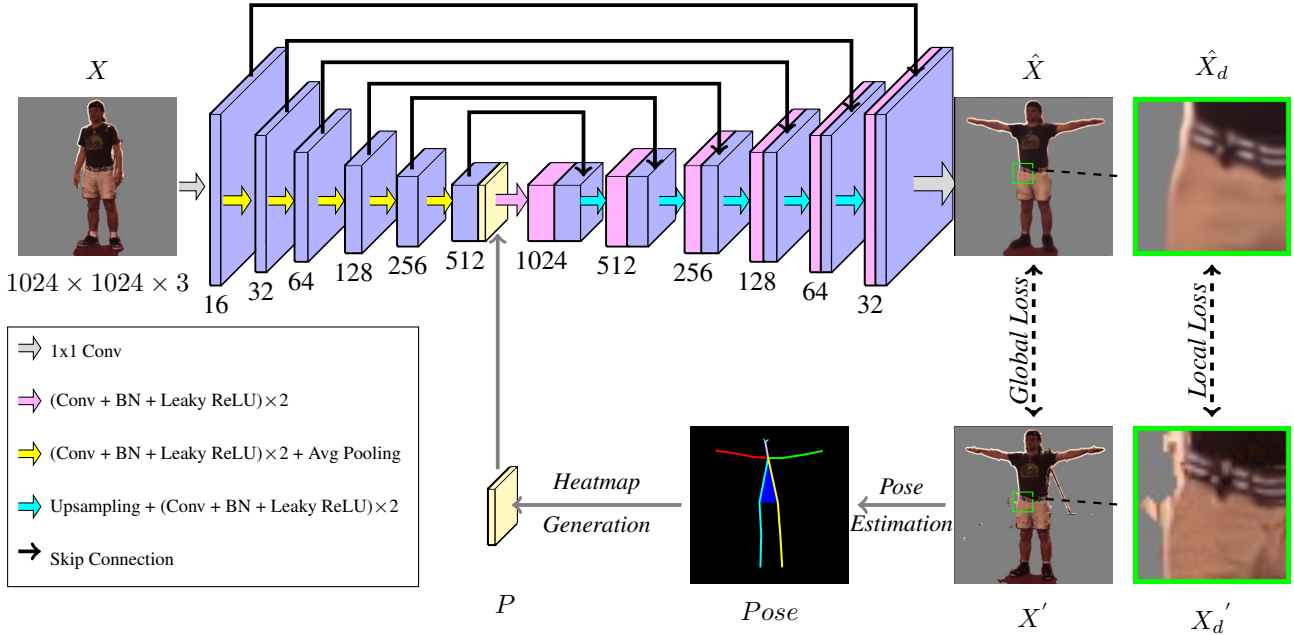}
    \caption{\textbf{Overall Network Architecture.} The reference image $X$ is first passed through an encoder to generate a latent representation. In the lower branch, 18 keypoints are estimated from the ground truth image $X^{'}$ to produce an explicit pose representation $P$. $P$ is then concatenated with the latent representation, which is further decoded into the output $\hat{X}$. Global perceptual loss is enforced between $X^{'}$ and $\hat{X}$. To enable high-resolution appearance transfer, two types of local loss are also enforced on the corresponding local regions (${X_d}^{'}$, $\hat{X_d}$), indicated by the bounding boxes. Local details are progressively refined. See section~\ref{sec:32} and section~\ref{sec:33} for technical contributions. }
    \label{fig:HPT}
\end{figure*}
\section{Related Work}
We address the problem of high-resolution pose-guided appearance transfer, which is also a problem of conditional image generation. Therefore, in this section, works related to conditional image generation and pose transfer will first be discussed, followed by recent approaches that can produce high-resolution images.\\[1ex]
\noindent\textbf{Conditional image generation} Generative models including Variational Autoencoders~\cite{VAE} (VAEs) and Generative Adversarial Networks~\cite{GAN} (GANs) have demonstrated success in image generation. Although VAEs can generate target images complying a given reference image, they may not faithfully preserve high-quality details because a lower bound is optimized.\par
Conditional GANs~\cite{CondGAN} have been exploited to solve many challenging tasks. Zhao \textit{et al.}~\cite{VariGAN} integrated GANs and other inference models to generate images of people in various clothing styles from multiple views. Reed \textit{et al.}~\cite{WhatWhere} proposed a conditional generative model that used pose and text as conditions to generate images. Lassner \textit{et al.}~\cite{GenerativeClothing} also presented a generative model that could generate realistic images conditioning on clothing segmentation.

Numerous researchers~\cite{Structured, CondSyn, CondVisual, StackGAN, FashionGAN} introduced their respective methods to enable more control on the appearance of the generated images in generative processes by providing different intermediate information such as labels and texts. Models such as Conditional GANs~\cite{CondGAN} and CycleGAN~\cite{CycleGAN} also demonstrated their efficacy in image-to-image translation.

However, it is difficult for the above methods to simultaneously encode different factors such as pose and appearance. To transfer the pose-invariant human appearance, disentangling pose and appearance from the reference image becomes an essential step. Many previous studies~\cite{InfoGAN, HiddenFactor} attempted to use GANs~\cite{GAN} and autoencoders~\cite{AE2011} to disentangle such factors, including writing styles from character identities. Recently, Tran \textit{et al.}~\cite{DRGAN} proposed DRGAN, which can disentangle pose from identity by learning the representation of human face followed by synthesizing the face with preserved identity at the target pose. \\[1ex]
\noindent\textbf{Appearance transfer\  } Recent work produced high-quality transfer results by employing 3D human models and information (in the data synthesis stage)~\cite{li2019dense} or by estimating 3D human model as an intermediate step~\cite{Zanfir_2018_CVPR}. With no 3D information used, approaches for pose transfer~\cite{DGPose, Disentangled2018} used encoder-decoders to attempt disentangling the pose and appearance of the input image to perform pose transfer. Esser \textit{et al.}~\cite{vunet2018} explored a variational U-Net~\cite{UNet} on transferring the pose of a reference image invariant with its appearance. \\
\indent The PG$^2$~\cite{2017PG2} was a more related work that aims at generating images of a subject in various poses based on an image of that person and one novel pose. Combining GANs and autoencoders, PG$^2$ was trained through an encoder-decoder network followed by a refinement network given the pose and person image as input. Siarohin \textit{et al.}~\cite{DSC} proposed a generative model similar to PG$^2$, where a discriminator was included at the end of the autoencoder to help generate realistic images. Instead of using a discriminator, the pose transfer network presented by Natalia \textit{et al.}~\cite{DensePose} attempted to produce the seamless result by blending the synthesized image and warped image through end-to-end training. Though not aiming at transferring human pose, the landmark learning network recently proposed by Jakab~\textit{et al.}~\cite{Jakab18} demonstrated acceptable results on pose transferring, which was achieved by using a simple encoder-decoder network with the learning landmarks concatenated in an intermediate representation.\\
\indent Although \cite{2017PG2, DensePose, Jakab18} performed well on changing pose at low-resolution ($128^2$) reference images while keeping their rough identity, they could not preserve but significantly blur complex textures after pose transfer. In contrast, we are dealing with a more challenging task compared to their work since we want to preserve as many details as possible at the target pose presented in the high-resolution reference image.\\[1ex]
\textbf{Progressive training\  } In the generative model, producing high-resolution and high-quality results is difficult since the training process becomes unstable and hard to converge as the output dimension increases. \\
\indent Recently, Tero~\textit{et al.}~\cite{ProgressiveGAN} proposed a progressive training methodology for GANs to generate high-quality results. They started training from low resolution and added layers to the model progressively to obtain satisfactory high-resolution results. Tero's work focused on GANs and cannot be directly applied to autoencoders while our goal is conditional image generation using autoencoders for even higher resolution output ($1024^2$).

\begin{figure*}
    \begin{center}
\vspace{-0.2in}
    \input{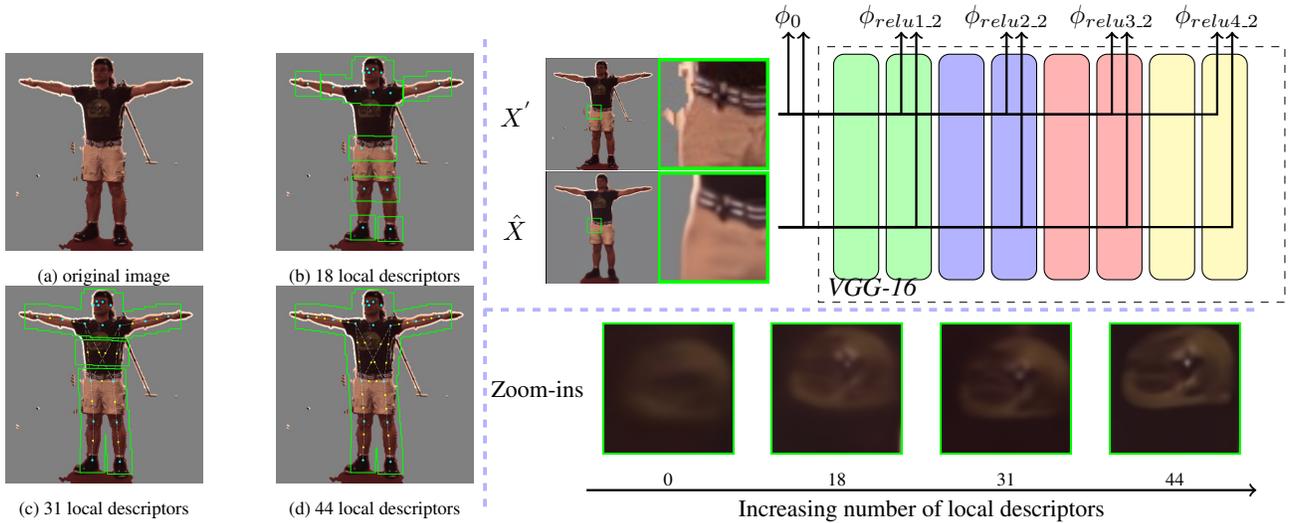}
    \end{center}
\vspace{-0.2in}
    \caption{\textbf{Left:} the distribution and coverage of different numbers of local descriptors. Local descriptors are centered at the dots and their coverage is indicated by green bounding boxes. In particular, blue dots denote the 18 keypoints generated by a pose estimator and yellow dots denote the interpolated keypoints. Denser local descriptors introduce higher coverage of human body. \textbf{Right-top:} the mechanism of local perceptual loss back-propagation. Two corresponding local regions $\hat{X_d}$ and $X_d'$ are respectively cropped from generated image $\hat{X}$ and ground truth image $X'$. $\hat{X_d}$ and ${X_d'}$ are then separately passed through a pre-trained VGG-16 to generate activations $\phi$ at different layers $l$. A customized criterion $C(\phi,{\phi'})$ measures the distances between corresponding activations $\phi$. Local descriptors intensify local loss back-propagation and thus enhance local details: see the sharper wrinkles and belt depicted in $X_d'$. \textbf{Right-bottom:} the detail enhancement introduced by increasingly denser overlapping local descriptors. Specifically, the four zoom-in views of logo are respectively cropped from outputs of four models trained using 0, 18, 31, 44 local descriptors (from left to right). Subtle but evident improvement can be identified in the process of increasing the number of overlapping local descriptors.}
    \label{fig:local}
\end{figure*}

\section{Method}
\vspace{-0.08in}
Figure~\ref{fig:HPT} details the autoencoder architecture for pose-guided appearance transfer with technical contributions to transfer appearance details at ${1024}^2$ to be detailed.\par
Specifically, given a reference image $X$ of a person and another image $X^{'}$ of the same person which is in the target pose, we first extract the explicit pose representation $P$ from $X^{'}$ using a state-of-the-art pose estimator (section~\ref{sec:31}). We then inject $P$ into the autoencoder's bottleneck by concatenating it with the deepest feature map generated by the encoder. Finally, the concatenated feature block is passed through a decoder to generate an image with the person in the target pose, denoted as $\hat{X}$. Reconstruction loss is enforced globally between $X^{'}$ and $\hat{X}$  (section~\ref{sec:31}). To enable high-resolution appearance transfer, we employ novel local descriptors to refine output details. Local descriptors are applied under the guidance of keypoint locations from the pose estimator (section~\ref{sec:32}). To generate images in a high resolution ($1024^2$), the encoder and decoder are grown progressively as training proceeds (section~\ref{sec:33}) and a super resolution (such as SRGAN~\cite{SRGAN2017}, SinGAN~\cite{shaham2019singan}) can be additionally applied to further sharpen simple garment textures.

\subsection{Pose-guided high-resolution appearance transfer}\label{sec:31}
\noindent\textbf{Pose representation} To represent human pose information in an explicit manner, we employ a state-of-the-art pose estimator~\cite{openpose}, which gives the locations of 18 keypoints of a person in 2D coordinates. To let the network leverage the keypoint information effectively, these 18 keypoints are separately represented by a gaussian distribution map with a fixed standard deviation. Specifically, we denote each keypoint as $k = 1, \cdots, 18$ and their respective 2D coordinates as $u(k)$. Then the pose representation $P$, which is the concatenation of 18 gaussian distribution maps, is encoded as:
\begin{equation}
    P(\textbf{x};k) = \exp(-\frac{1}{2\sigma^2}{||\textbf{x} - u(k)||}^2)
\end{equation}
The result is an explicit pose representation $P\in\mathbb{R}^{H\times W \times 18}$ whose 18 maxima represent the locations of the 18 keypoints. $P$ is then concatenated into the bottleneck of autoencoder.\\ [1ex]
\textbf{Autoencoder} The goal of the autoencoder is to reconstruct $\hat{X}$ in the target pose based on the appearance of the person in the reference image $X$ and the pose representation $P$ extracted from the same person in the ground truth image $X^{'}$, as shown in Figure~\ref{fig:HPT}. Since $P$ contains no appearance information, the network is forced to utilize the appearance information in $X$. Furthermore, we add skip connections similar to those in a U-Net ~\cite{UNet} to enable smoother gradient flow along the autoencoder. Then we use reconstruction loss between output $\hat{X}$ and ground truth image $X^{'}$ to encourage the network to generate appropriate appearance which matches the pose of the person in $X^{'}$.\\ [1ex]
\textbf{Perceptual loss} The design of reconstruction error is critical for good performance. Since it is hard for the network to learn a pixel-to-pixel mapping only from $X$ due to the inherent pose and appearance variation, we encourage the network to also learn high-level semantic meanings during training, which is pivotal for decoupling pose and appearance. Inspired by recent excellent practices~\cite{PerceptualLoss}, we adopt perceptual loss as the reconstruction loss between $X^{'}$ and $\hat{X}$. Apart from comparing only the raw pixel values, perceptual loss involves passing the output and the ground truth images individually through a pre-trained deep network and comparing the activations extracted from multiple layers inside the network. This process enables the network to better learn the decoupling of appearance and pose and alleviates overfitting. Specifically, we define perceptual loss as:
\begin{equation}
    L(X^{'},\hat{X})= \sum_{l} C({\phi}_{l}(X^{'}),{\phi}_{l}(\hat{X}))
\end{equation}
where $\phi(x)$ is a pre-trained network, such as VGG-16~\cite{VGG16}, and ${\phi}_l$ denotes the activation of the $l^{th}$ layer of $\phi(x)$. Different from common practices which use $L_2$ loss as the criterion to evaluate $\hat{X}$, we customize the criterion $C(\phi, {\phi}^{'})$ to accelerate network convergence. Since $L_2$ loss has an optimal solution while $L_1$ loss enforces sharper output but is less stable, we designate $C(\phi, {\phi}^{'})$ as $L_2$ loss in the first half of the training process within each resolution level and $L_1$ loss in the second half. This practice enables stable convergence as well as high generation quality.
\begin{figure}
\begin{center}
\begin{adjustbox}{max width=0.7\linewidth}
\begin{tikzpicture}[scale = 0.9]

\def \imageboxwidth {0.12\linewidth};
\def \imagewidth {0.12\linewidth};
\def \textheight {0.6cm};
\def \targettop {0.31835938}
\def \targetleft {0.18554688}


\def \imageyoffset {1.5cm}


\newcommand{\networkLayer}[8]{
    \def\a{#1} 
    \def\b{0.02}
    \def\c{#2} 
    \def\t{#3} 
    \def\d{#4} 

    \draw[line width=0.3mm](\c+\t,0,\d) -- (\c+\t,\a,\d) -- (\t,\a,\d);                                                      
    \draw[line width=0.3mm](\t,0,\a+\d) -- (\c+\t,0,\a+\d) node[midway,below] {#6} -- (\c+\t,\a,\a+\d) -- (\t,\a,\a+\d) -- (\t,0,\a+\d) node[midway,left] {#8}; 
    \draw[line width=0.3mm](\c+\t,0,\d) -- (\c+\t,0,\a+\d);
    \draw[line width=0.3mm](\c+\t,\a,\d) -- (\c+\t,\a,\a+\d);
    \draw[line width=0.3mm](\t,\a,\d) -- (\t,\a,\a+\d) node[midway,above,yshift=2 mm] {#7};

    \filldraw[#5] (\t+\b,\b,\a+\d) -- (\c+\t-\b,\b,\a+\d) -- (\c+\t-\b,\a-\b,\a+\d) -- (\t+\b,\a-\b,\a+\d) -- (\t+\b,\b,\a+\d); 
    \filldraw[#5] (\t+\b,\a,\a-\b+\d) -- (\c+\t-\b,\a,\a-\b+\d) -- (\c+\t-\b,\a,\b+\d) -- (\t+\b,\a,\b+\d);

    \ifthenelse {\equal{#5} {}}
    {} 
    {\filldraw[#5] (\c+\t,\b,\a-\b+\d) -- (\c+\t,\b,\b+\d) -- (\c+\t,\a-\b,\b+\d) -- (\c+\t,\a-\b,\a-\b+\d);} 
}

\newcommand{\networkArrow}[4]{
    \def\xoffset{#1}
    \def\yoffset{#2}
    \def\length{#3}
    \draw[fill=#4](\xoffset-\length-\length-\length,2mm+\yoffset)--(\xoffset-\length,2mm+\yoffset)--(\xoffset-\length,2mm+\yoffset-\length)--(\xoffset,2mm+\yoffset+0.5*\length)--(\xoffset-\length,2mm+\yoffset+\length+\length)--(\xoffset-\length,2mm+\yoffset+\length)--(\xoffset-\length-\length-\length,2mm+\yoffset+\length);
}

\newcommand{\networkLine}[4]{
    \def\startx{#1}
    \def\starty{#2}
    \def\endx{#3}
    \def\endy{#4}
    \draw[->,line width=0.5mm] (\startx,\starty)--(\startx,\endy)--(\endx,\endy)->(\endx,\starty)

}

\def \initw {2pt};
\def \initd {0.12pt};
\def \spacing {0.4pt};
\def \ratioD {5};
\def \ratioW {1/5};
\def \ratiod {5/4};
\def \ratiow {4/5};

\setlength{\offset}{0.2pt}
\setlength{\ww}{\initw*\ratiow}
\setlength{\dd}{\initd}
\setlength{\arrowoffset}{3mm}

\definecolor{color1}{RGB}{255,0,255}
\definecolor{color2}{RGB}{255,255,0}
\definecolor{color3}{RGB}{0,255,255}


\networkLayer{\ww}{\dd}{0.0}{0.0}{color=white!30}{3}{64}{64};

\networkArrow{\arrowoffset}{2mm+1mm}{2mm}{gray!30};

\setlength{\offset}{\offset+\spacing+\spacing}
\setlength{\dd}{\dd*\ratiod}
\networkLayer{\ww}{\dd}{\offset}{0.0}{color=blue!30}{256}{}{};

\setlength{\offset}{\offset+\spacing+\spacing}
\setlength{\ww}{\ww*\ratiow*\ratiow}
\setlength{\dd}{\dd*\ratiod*\ratiod}

\networkLayer{\ww}{\dd}{\offset}{0.0}{color=blue!30}{512}{}{};
\networkLayer{\ww}{\dd*\ratiow*\ratiow}{\offset+\dd}{0.0}{color=yellow!30}{}{}{};
\draw[line width = 0.2mm,->,dashed](\arrowoffset+16mm,-9mm)--(\arrowoffset+16mm,-4mm)node[below,yshift=-4mm,xshift=3mm]{\small{pose injection}};
\setlength{\arrowoffset}{\arrowoffset+10.5mm}
\networkArrow{\arrowoffset}{2mm}{2mm}{color2};
\setlength{\offset}{\offset+\spacing+\spacing+\dd}
\networkLayer{\ww}{\dd}{\offset}{0.0}{color=color1!30}{1024}{}{};
\networkLayer{\ww}{\dd}{\offset+\dd}{0.0}{color=blue!30}{}{}{};

\setlength{\arrowoffset}{\arrowoffset+11mm}
\networkArrow{\arrowoffset}{2mm}{1.6mm}{color1!30};
\setlength{\offset}{\offset+\spacing+\spacing+\dd+\dd}
\setlength{\ww}{\ww*\ratiod*\ratiod}
\setlength{\dd}{\dd*\ratiow*\ratiow}
\networkLayer{\ww}{\dd}{\offset}{0.0}{color=color1!30}{512}{}{};
\networkLayer{\ww}{\dd}{\offset+\dd}{0.0}{color=blue!30}{}{}{};

\setlength{\arrowoffset}{\arrowoffset+10mm}
\networkArrow{\arrowoffset}{2mm}{1.6mm}{color3};
\setlength{\offset}{\offset+\spacing+\spacing+\dd+\dd}
\setlength{\dd}{\dd*\ratiow}
\networkLayer{\ww}{\dd}{\offset}{0.0}{color=white!30}{3}{}{};
\setlength{\arrowoffset}{\arrowoffset+11mm}
\networkArrow{\arrowoffset}{2mm+1mm}{2mm}{gray!30};


\setlength{\offset}{3pt}
\setlength{\ww}{\initw}
\setlength{\dd}{\initd}
\setlength{\arrowoffset}{-8.5mm}
\def\arrowyoffset{-38mm}

\networkLayer{\ww}{\dd}{\offset}{10}{color=white!30}{3}{128}{128};

\networkArrow{\arrowoffset}{\arrowyoffset+4mm}{2mm}{gray!30};

\setlength{\offset}{\offset+\spacing+\spacing}
\setlength{\dd}{\dd*\ratiod}
\networkLayer{\ww}{\dd}{\offset}{10}{color=blue!30}{128}{}{};

\setlength{\arrowoffset}{\arrowoffset+9mm}
\networkArrow{\arrowoffset}{\arrowyoffset}{2mm}{color2};
\setlength{\offset}{\offset+\spacing+\spacing}
\setlength{\ww}{\ww*\ratiow}
\setlength{\dd}{\dd*\ratiod}

\networkLayer{\ww}{\dd}{\offset}{10}{color=blue!30}{256}{}{};

\setlength{\arrowoffset}{\arrowoffset+11mm}
\networkArrow{\arrowoffset}{\arrowyoffset}{2mm}{color2};
\setlength{\offset}{\offset+\spacing+\spacing}
\setlength{\ww}{\ww*\ratiow*\ratiow}
\setlength{\dd}{\dd*\ratiod}

\networkLayer{\ww}{\dd}{\offset}{10}{color=blue!30}{512}{}{};
\networkLayer{\ww}{\dd*\ratiow}{\offset+\dd}{10}{color=yellow!30}{}{}{};
\setlength{\arrowoffset}{\arrowoffset+10mm}
\networkArrow{\arrowoffset}{\arrowyoffset}{1.6mm}{color1!30};
\draw[line width = 0.2mm,->,dashed](\arrowoffset-5mm,\arrowyoffset-10mm)--(\arrowoffset-5mm,\arrowyoffset-5mm)node[below,yshift=-4mm,xshift=3mm]{\small{pose injection}};
\setlength{\offset}{\offset+\spacing+\spacing+\dd}
\networkLayer{\ww}{\dd}{\offset}{10}{color=color1!30}{1024}{}{};
\networkLayer{\ww}{\dd}{\offset+\dd}{10}{color=blue!30}{}{}{};

\setlength{\arrowoffset}{\arrowoffset+10.5mm}
\networkArrow{\arrowoffset}{\arrowyoffset}{1.6mm}{color3};
\setlength{\offset}{\offset+\spacing+\spacing+\dd+\dd}
\setlength{\ww}{\ww*\ratiod*\ratiod}
\setlength{\dd}{\dd*\ratiow}
\networkLayer{\ww}{\dd}{\offset}{10}{color=color1!30}{512}{}{};
\networkLayer{\ww}{\dd}{\offset+\dd}{10}{color=blue!30}{}{}{};

\setlength{\arrowoffset}{\arrowoffset+10.5mm}
\networkArrow{\arrowoffset}{\arrowyoffset}{1.8mm}{color3};

\setlength{\offset}{\offset+\spacing+\spacing+\dd+\dd}
\setlength{\ww}{\ww*\ratiod}
\setlength{\dd}{\dd*\ratiow}
\networkLayer{\ww}{\dd}{\offset}{10}{color=color1!30}{256}{}{};
\networkLayer{\ww}{\dd}{\offset+\dd}{10}{color=blue!30}{}{}{};
\setlength{\arrowoffset}{\arrowoffset+10.5mm}
\networkArrow{\arrowoffset}{\arrowyoffset+4mm}{2mm}{gray!30};
\setlength{\offset}{\offset+\spacing+\spacing+\dd+\dd}
\setlength{\dd}{\dd*\ratiow}
\networkLayer{\ww}{\dd}{\offset}{10}{color=white!30}{3}{}{};

\networkLine{7mm}{12mm}{40mm}{20mm};
\networkLine{17mm}{8mm}{30mm}{16mm};

\networkLine{-4mm}{-23mm}{48mm}{-15mm};
\networkLine{5mm}{-26mm}{38mm}{-18mm};
\networkLine{14mm}{-30mm}{28mm}{-24mm};

\end{tikzpicture}
\end{adjustbox}
\end{center}
\vspace{-0.2in}
  \caption{\textbf{Progressive training.} The bottleneck size of the autoencoder is $32 \times 32$. We start from a low spacial resolution of $64 \times 64$ pixels and incrementally add layers to encoder and decoder as training proceeds until we reach the ultimate resolution of $1024 \times 1024$. All existing layers remain trainable throughout the process. Here we illustrate a snapshot when the network increases its resolution from $64 \times 64$ to $128 \times 128$. During this transition, a new convolution block [(Conv + BN + Leaky ReLU) $\times$ 2] with corresponding up-sampling or down-sampling layer is introduced to encoder and decoder respectively. Also $1\times1$ convolution layer used to project RGB channels to/from feature space is replaced by a new one that fits the network.}
\vspace{-0.2in}

  \label{fig:Progressive}
\end{figure}
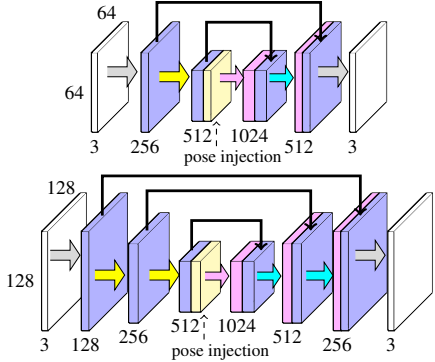
\subsection{Local descriptors}\label{sec:32}
Note that the adoption of global perceptual loss does not enforce sufficient preservation of local details. It is observed that sharp garment textures cannot be preserved well under the restriction of global perceptual loss only, as will be shown in the ablation study in Figure~\ref{fig:self_comparison}. To address this limitation, we introduce novel local descriptors which enable the generation of high-quality images. Local descriptors describe a set of regions telling the network where to focus and concentrate loss back-propagation. The locations of local descriptors are guided by the pose keypoints produced by the pose estimator. To ensure appropriate detail refinement and alleviate overfitting, the size of local regions is designed to be one-eighth of the input image resolution.\par
Figure~\ref{fig:local} shows the distribution and coverage of local descriptors. Since higher resolution generally requires more local details, we increase the number of local descriptors adopted by interpolating between existing keypoints as input image resolution grows. Denser overlapping local descriptors introduce more complete coverage of the body and thus help preserve details more faithfully. Since we are not interested in fingers so we do not incorporate the keypoints particularly there.\par
Two kinds of local descriptor are adopted, respectively local perceptual loss and local discriminator. We use local perceptual loss as local descriptors during progressive training and local discriminators at the highest resolution ($1024^2$).\par
\begin{figure}
  \begin{tikzpicture}[font=\tiny, scale = 0.5]

\def \imageboxwidth {0.09\linewidth};
\def \imagewidth {0.075\linewidth};
\def \textheight {0.6cm};


\def \imageroot {images/local_D_structure/2};

\def \imageyoffset {1.5cm}
%
%

\def \targettop {0.32617188}
\def \targetleft {0.1953125}

\node[label=left:{(a)${X_d}$}] (target_patch) at (-3.5cm, \imageboxwidth/2+2*\textheight+\imageyoffset)
    {\adjincludegraphics[width=\imagewidth]{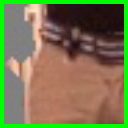}};

\def \blendtop {0.32421875}
\def \blendleft {0.17773438}

\node[label={[xshift = -7.8mm,yshift=-6.8mm](b)${X_d}^{'}$}] (blend_patch) at (-3.5cm, -\imageboxwidth/2+\imageyoffset)
    {\adjincludegraphics[width=\imagewidth]{images/local/gt_local_1.png}};

\node[label=left:{(c)$\hat{X}_{d}$}] (transfer_patch) at (-3.5cm, -\imageboxwidth*3+\imageyoffset)
    {\adjincludegraphics[width=\imagewidth]{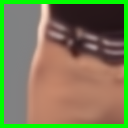}};

\draw[green, very thick, ->] (target_patch.east) -- (-2.3cm, 1.5cm) -- (-1.7cm, 1.5cm);
\draw[green, very thick, ->] (blend_patch.east) -- (-2.3cm, 1.5cm) -- (-1.7cm, 1.5cm);

\draw[red, very thick, ->] (target_patch.east) -- (-2.3cm, -0.5cm) -- (-1.7cm, -0.5cm);
\draw[red, very thick, ->] (transfer_patch.east) -- (-2.3cm, -0.5cm) -- (-1.7cm, -0.5cm);



\newcommand{\networkLayer}[8]{
	\def\a{#1} 
	\def\b{0.02}
	\def\c{#2} 
	\def\t{#3} 
	\def\d{#4} 

	\draw[line width=0.3mm](\c+\t,0,\d) -- (\c+\t,\a,\d) -- (\t,\a,\d);                                                      
	\draw[line width=0.3mm](\t,0,\a+\d) -- (\c+\t,0,\a+\d) node[midway,below] {#6} -- (\c+\t,\a,\a+\d) -- (\t,\a,\a+\d) -- (\t,0,\a+\d) node[midway,xshift=-0.2cm] {#8}; 
	\draw[line width=0.3mm](\c+\t,0,\d) -- (\c+\t,0,\a+\d);
	\draw[line width=0.3mm](\c+\t,\a,\d) -- (\c+\t,\a,\a+\d);
	\draw[line width=0.3mm](\t,\a,\d) -- (\t,\a,\a+\d) node[midway,above,xshift=0.04cm,yshift=0.05cm] {#7};

	\filldraw[#5] (\t+\b,\b,\a+\d) -- (\c+\t-\b,\b,\a+\d) -- (\c+\t-\b,\a-\b,\a+\d) -- (\t+\b,\a-\b,\a+\d) -- (\t+\b,\b,\a+\d); 
	\filldraw[#5] (\t+\b,\a,\a-\b+\d) -- (\c+\t-\b,\a,\a-\b+\d) -- (\c+\t-\b,\a,\b+\d) -- (\t+\b,\a,\b+\d);

	\ifthenelse {\equal{#5} {}}
	{} 
	{\filldraw[#5] (\c+\t,\b,\a-\b+\d) -- (\c+\t,\b,\b+\d) -- (\c+\t,\a-\b,\b+\d) -- (\c+\t,\a-\b,\a-\b+\d);} 
}


\def \initw {4.0pt};
\def \initd {0.2pt};
\def \spacing {0.33pt};
\def \ratiod {3/2};
\def \ratiow {2/3};


\networkLayer{\initw}{0.1}{0.0}{0.0}{color=white!30}{6}{128}{128};


\setlength{\offset}{0.5pt}
\setlength{\ww}{\initw*\ratiow}
\setlength{\dd}{\initd}

\networkLayer{\ww}{\dd}{\offset}{0.0}{color=blue!30}{64}{64}{};
\networkLayer{\ww}{0.1}{\offset+\dd+0.1}{0.0}{color=red!30}{}{}{};
\networkLayer{\ww}{0.1}{\offset+\dd+0.3}{0.0}{color=yellow}{}{}{};


\setlength{\offset}{\offset+\dd+0.3pt+\spacing}
\setlength{\ww}{\ww*\ratiow}
\setlength{\dd}{\dd*\ratiod}

\networkLayer{\ww}{\dd}{\offset}{0.0}{color=blue!30}{128}{32}{};
\networkLayer{\ww}{0.1}{\offset+\dd+0.1}{0.0}{color=red!30}{}{}{};
\networkLayer{\ww}{0.1}{\offset+\dd+0.3}{0.0}{color=yellow}{}{}{};


\setlength{\offset}{\offset+\dd+0.3pt+\spacing}
\setlength{\ww}{\ww*\ratiow}
\setlength{\dd}{\dd*\ratiod}

\networkLayer{\ww}{\dd}{\offset}{0.0}{color=blue!30}{256}{16}{};
\networkLayer{\ww}{0.1}{\offset+\dd+0.1}{0.0}{color=red!30}{}{}{};
\networkLayer{\ww}{0.1}{\offset+\dd+0.3}{0.0}{color=yellow}{}{}{};


\setlength{\offset}{\offset+\dd+0.3pt+\spacing}
\setlength{\ww}{\ww*\ratiow}
\setlength{\dd}{\dd*\ratiod}

\networkLayer{\ww}{\dd}{\offset}{0.0}{color=blue!30}{512}{8}{};
\networkLayer{\ww}{0.1}{\offset+\dd+0.1}{0.0}{color=red!30}{}{}{};
\networkLayer{\ww}{0.1}{\offset+\dd+0.3}{0.0}{color=yellow}{}{}{};


\setlength{\offset}{\offset+\dd+0.3pt+\spacing}
\setlength{\ww}{\ww*\ratiow}
\setlength{\dd}{\dd*\ratiod}

\networkLayer{\ww}{\dd}{\offset}{0.0}{color=blue!30}{1024}{4}{};
\networkLayer{\ww}{0.1}{\offset+\dd+0.1}{0.0}{color=red!30}{}{}{};
\networkLayer{\ww}{0.1}{\offset+\dd+0.3}{0.0}{color=yellow}{}{}{};


\setlength{\offset}{\offset+\dd+0.3pt+\spacing}
\setlength{\ww}{\ww*\ratiow}
\setlength{\dd}{\dd*\ratiod}

\networkLayer{\ww}{\dd}{\offset}{0.0}{color=blue!30}{2048}{2}{};
\networkLayer{\ww}{0.1}{\offset+\dd+0.1}{0.0}{color=red!30}{}{}{};
\networkLayer{\ww}{0.1}{\offset+\dd+0.3}{0.0}{color=yellow}{}{}{};


\setlength{\offset}{\offset+\dd+0.3pt+\spacing+0.2pt}

\networkLayer{\ww}{0.1}{\offset}{0.0}{color=blue!30}{1}{2}{};
\networkLayer{\ww}{0.1}{\offset+0.1+0.1}{0.0}{color=red!30}{}{}{};
\networkLayer{\ww}{0.1}{\offset+0.2+0.1+0.1}{0.0}{color=orange!30}{}{}{};
\networkLayer{0.1}{0.1}{\offset+0.4+\spacing}{0.0}{color=pink!30}{1}{}{};


\node(positive_D) at (10.15cm, 1.5cm) {$D_{local}(X_d, {X_d}^{'})\rightarrow \textcolor{green}{1}$};
\node(negative_D) at (10.15cm, -1.5cm) {$D_{local}(X_d, \hat{X}_{d})\rightarrow \textcolor{red}{0}$};

\draw[->, green, very thick] (9.55cm, 0cm) -- (positive_D.south);
\draw[->, red, very thick] (9.55cm, 0cm) -- (negative_D.north);


\matrix [draw,below left] at ($(current bounding box.north)+(8cm, 1cm)$) {
  \node [draw,minimum width=1mm,minimum height=1mm,fill=blue!30, label=right:{Conv3x3 (padding=2)}] {}; &
  \node [draw,minimum width=1mm,minimum height=1mm,fill=yellow, label=right:{Leaky ReLU}] {}; \\
  \node [draw,minimum width=1mm,minimum height=1mm,fill=red!30, label=right:{Spectral Normalization}] {}; &
  \node [draw,minimum width=1mm,minimum height=1mm,fill=orange!30, label=right:{Sigmoid}] {}; \\
  \node [draw,minimum width=1mm,minimum height=1mm,fill=pink!30, label=right:{Avg Pooling}] {}; & \\
};

\end{tikzpicture}
\vspace{-0.2in}
  \caption{\textbf{Local Discriminators.} ${X_d}$, ${X_d}^{'}$ and $\hat{X_d}$ are cropped from the reference image, the target image (GT) and the generated image respectively. The concatenation of ${X_d}$ and ${X_d}^{'}$ is fed into the local discriminator as a positive example, while the concatenation of ${X_d}^{'}$ and $\hat{X_d}$ is fed as a negative example. Shown here is the architecture of the local discriminator with the last 2x2x1 feature vector being averaged to a scalar as the probability of the input pair being a valid transfer result.}
\vspace{-0.2in}
  \label{fig:local_discriminator}
\end{figure}
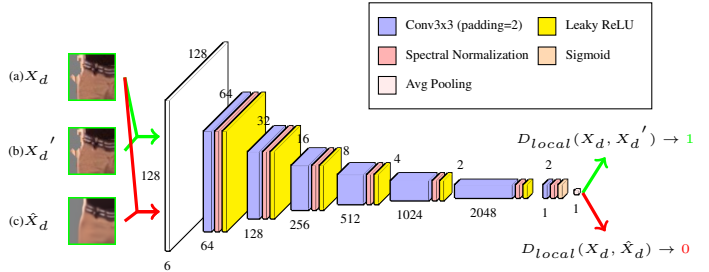

Specifically, based on the 18 keypoints in $X^{'}$ produced by the pose estimator, a list of $N$ local descriptors is generated, denoted as $d = 1, \cdots, N$. Then two sets of fractional-sized regions centered at the location of each of $N$ local descriptors are cropped from $X^{'}$ and $\hat{X}$ respectively. \\ [1ex]
\noindent\textbf{Local perceptual loss} Perceptual loss is enforced between corresponding local regions. The local loss $L_{local}$ is formulated as the following:
\begin{equation}
    L_{local}(X^{'}, \hat{X}) = \sum_{d=1}^{N}\sum_{l}C({\phi}_{l}(X_d^{'}), {\phi}_{l}(\hat{X}_d)))
\end{equation}
where ${X_d}^{'}$ and ${\hat{X}}_d$ denote the $d^{th}$ region cropped from $X^{'}$ and $\hat{X}$ respectively. \\ [1ex]
\noindent\textbf{Local discriminators} We adopt local discriminators in replacement of the local perceptual loss at the highest resolution ($1024^2$). Specifically, the local discriminators at the $d^{th}$ region take pairs of inputs where the concatenation of ${X_d}$ and ${X_d}^{'}$ is considered real and the concatenation of ${X_d}$ and ${\hat{X}}_d$ is considered fake, as shown in Figure~\ref{fig:local_discriminator}. It is observed that local discriminators can improve the generalization ability of the model and further boost its generation quality.\par
Self-comparison between the model with and without local descriptors are shown in Figure~\ref{fig:self_comparison}. The significant improvement in image quality demonstrates the efficacy introduced by local descriptors.

\subsection{Progressive training of autoencoder}\label{sec:33}
\begin{figure*}[t]
  \input{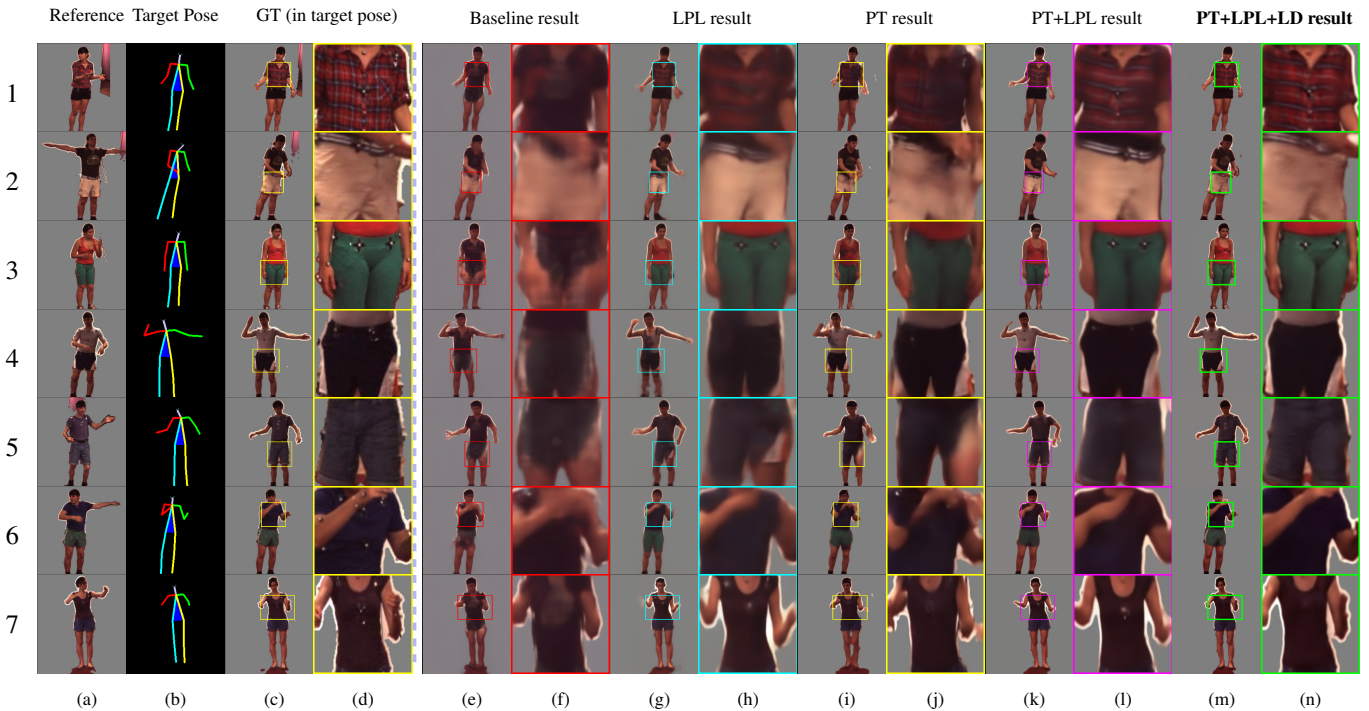}
  \vspace{-0.2in}
  \caption{\textbf{Self-comparison results.} Test results on the Human3.6M generated by Baseline (no local descriptors or progressive training), LPL (with local perceptual loss only), PT (with progressive training), PT+LPL (with progressive training and local perceptual loss), PT+LPL+LD (with progressive trainning, local perceptual loss and local discriminators) and their corresponding zoom-in views are provided. Progressive training and local descriptors (local perceptual loss + local discriminator) each introduces considerable improvement in generation quality and produce the best result when combined. Our model also demonstrates robustness to the segmentation error introduced by the Human3.6M dataset.  Figure is best viewed online.}
\vspace{-0.2in}
  \label{fig:self_comparison}
\end{figure*}

Apart from achieving high-quality image generation, we also aim at producing unprecedentedly high-resolution results ($1024^2$). However, training the autoencoder in high resolution from scratch does not yield satisfactory results. Inspired by ~\cite{ProgressiveAE} which produces high-resolution results on CelebA-HQ dataset by introducing progressive training to GAN, we adopt a variation of progressive training which fits our setting of autoencoder with skip connections, as shown in Figure~\ref{fig:Progressive}. Most importantly, instead of fading in a new convolution block to increase resolution using alpha blending, we train the new convolution block with skip connection from scratch, utilizing deeper convolution blocks trained in the previous stage as mature feature extractors. From our observation, this enables faster convergence of newly introduced blocks as well as utilization of skip connections to enhance generation quality. Self-comparison in Figure~\ref{fig:self_comparison} demonstrates substantial improvement brought by progressive training on autoencoder.

\subsection{Implementation details}
We use the Adam~\cite{Adam} optimizer with a weight decay of $5\times10^{-4}$. The initial learning rate is set to $2 \times 10^{-4}$. We use $\sigma=3.2$ to generate the gaussian distribution for pose representation. The autoencoder is trained progressively starting from the resolution of $64^2$ with bottleneck shape of $1024 \times 32^2$ and ending at the resolution of $1024^2$. Within each convolution block, we use two contiguous sets of $3 \times 3$ convolution layer followed by batch normalization ~\cite{BatchNorm} and leaky Relu with leakiness of $0.2$. The number of channels of feature maps is halved as spacial size doubles. We downscale and upscale the feature maps using average pooling and nearest neighbor interpolation respectively. We use ${1}\times{1}$ convolution to project the outermost feature maps into RGB space and vice versa as in RGB back to feature map. We use He's initializer~\cite{HeInit} to initialize the autoencoder. A total of 18 local descriptors are used for the resolution of $64^2$ and $128^2$. For $256^2$ and $512^2$, we use 31 local descriptors by interpolating between keypoints pairs and 44 local descriptors for $1024^2$ through additional interpolations. For each resolution level, we train the network for $700$ thousand iterations.\par
Our final loss $L$, which is composed of both global loss $L_{global}$ and local loss $L_{local}$, is formulated as the following:
\begin{align}
        L(X^{'}, \hat{X}) &= L_{global}(X^{'}, \hat{X}) + L_{local}(X^{'}, \hat{X}) \nonumber \\ &= \sum_{l}C({\phi}_{l}(X^{'}), {\phi}_{l}(\hat{X}))) \nonumber \\  &+\sum_{d=1}^{N}\sum_{l}C({\phi}_{l}(X_d^{'}), {\phi}_{l}(\hat{X}_d))
\end{align}
\section{Experiments}

To demonstrate the advantages of our method, we first conduct qualitative and quantitative self-comparisons to validate the effectiveness of different components, namely local descriptors and progressive training on autoencoders. We then demonstrate the network's generalizability by showing the results on various datasets, including the Human3.6M~\cite{Ionescu2014Human36MLS} dataset, the DeepFashion~\cite{Liu2016DeepFashionPR} dataset and a self-collected dataset from YouTube. We also compare the performance on the DeepFashion dataset with previous work. Lastly, we show the network's potential to be further utilized in real-world applications, such as high-quality garment transfer and pose-guided human video generation.\par
\subsection{Datasets}
\noindent\textbf{Human3.6M} We train and test our model primarily on the Human3.6M
dataset~\cite{Ionescu2014Human36MLS}, which includes 11 actors in total with different poses. The dataset provides ground truth 2D human poses,
backgrounds and human body bounding boxes. We first subsample the videos at 3 frames per second and obtain image frames with large pose variations. For each image frame, we then subtract the background and retain only the human foreground to reduce training noises. We select `Posing', `Greeting' and 'Walking' action classes for training, and `Directions' class for testing.\par
\noindent\textbf{YouTube dataset} To test the generalizability of our method, we further train and test our network on our self-collected YouTube video datasets. The datasets we collected contains 20 hip-hop dancing videos from World of Dance competition, all of which have large pose variations. We subtract the background of this dataset using human parsing network~\cite{Liang2018LookIP} and subsample the videos at 3 frames per second to produce the training and testing set.

\begin{table}[t]
  \begin{center}
    \caption{Quantitative self-comparison in different modes.}
\vspace{-0.15in}
\begin{adjustbox}{max width=0.6\linewidth}
    \label{tab:self_comparison}
    \begin{tabular}{l|c c c} 
  \hline
  \multicolumn{4}{c}{Human3.6M}\\
  \textbf{Model} & \textbf{SSIM} & \textbf{local-SSIM} &\textbf{LPIPS}\\
  \hline
  Baseline & 0.909 & 0.699 & 0.230\\
  LPL & 0.944 & 0.744 &0.205\\
  PT & 0.953 & 0.759 &0.164\\
  PT+LPL & 0.954 & 0.772 &0.145\\
  PT+LPL+LD & \textbf{0.959} & \textbf{0.804} &\textbf{0.135}\\
  \hline
  Real Data & 1.00 & 1.00 &0.00
\end{tabular}

\end{adjustbox}
\vspace{-0.25in}
  \end{center}
\vspace{-0.15in}
\end{table}

\subsection{Self-comparison}
\noindent\textbf{Local descriptors}
Qualitative comparison in Figure~\ref{fig:self_comparison} demonstrates the effectiveness of local descriptors (local perceptual loss and local discriminators). As shown in column (e), (g) and their corresponding zoom-in views, local perceptual loss results in improvement on local details compared to the baseline. From column (i), (k) and their corresponding zoom-in views, local perceptual loss is still able to bring significant enhancement to the generation quality under progressive training. In particular, the two stars in image (3,~\textit{d}) are faithfully preserved in result (3,~\textit{l}), but lost in result (3,~\textit{j}). In addition, column (k), (m) and their corresponding zoom-in views show the further enhancement on the generation quality of human body and garment textures introduced by the local discriminators applied at the highest resolution ($1024^2$).\par
\noindent\textbf{Progressive training}
The advantages of progressive training is demonstrated through comparisons in Figure~\ref{fig:self_comparison}. Column (g) and (k) with their corresponding zoom-in views show the improvement for the models with local descriptors, while column (e) and (i) with zoom-in views show the improvement without local descriptors. In particular, the garment texture in image (1, \textit{d}) is faithfully preserved in result (1, \textit{l}), but lost in result (1, \textit{h}). Therefore, while local descriptors enable local detail enhancement, the entire network still suffers from the vanishing gradient problem. To alleviate this problem, progressive training enables separate and progressive convergence of deep network layers, thus reduce the effects of vanishing gradient.
\par
\noindent\textbf{Quantitative comparison}
Image generation quality can be difficult to assess due to various standards. Here we adopt Structural Similarity (SSIM)~\cite{wang2004SSIM} and Learned Perceptual Image Patch Similarity (LPIPS)~\cite{zhang2018perceptual} as our main evaluation metric. Due to the limitations of SSIM such as insensitivity and distortion under-estimation near hard edges~\cite{Pambrun2015Limitations}, we also adopt a variation of SSIM, local-SSIM, to more effectively evaluate local details. Instead of global evaluation performed by SSIM, local-SSIM operates on 44 corresponding local regions between the generated image and the reference image. The 44 local regions correspond to the areas described by 44 local descriptors, where the highest coverage of human body is achieved.\par
Quantitative comparison between models under different settings are shown in Table~\ref{tab:self_comparison}. Both local descriptors and progressive training bring considerable enhancement in generation quality, with a combination of the two further boosting the result. Local-SSIM more evidently reflects the improvement in the quality at local regions.

\subsection{YouTube dataset}
Since appearance variation of the Human3.6M dataset is extremely limited (\textit{i.e.} many actors wear garment with plain texture), we train and test our model on our YouTube dataset to validate its generalizability. Test results are shown in Figure~\ref{fig:youtube}. Our method clearly demonstrates its power in detail preservation and pose manipulation even under hard situations where large pose variations introduce different self-occlusions and thus dis-occlusion.

\subsection{Comparison with previous work}

\begin{figure}[t]
    \input{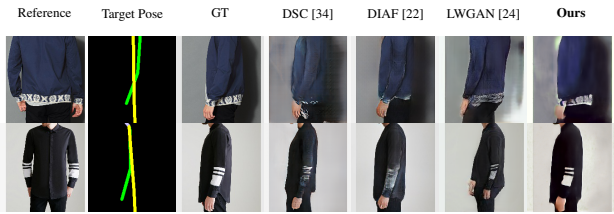}

 \vspace{-0.2in}
    \caption{\textbf{Comparison with previous work.} Comparing with previous work, given the large pose differences our network captures more details despite that some parts may not be sharp enough.}
    \label{fig:sota_comparison}
\vspace{-0.1in}
\end{figure}

Here we compare our method with three state-of-the-art approaches, DSC proposed by Siarohin \textit{et al.}, Dense Intrinsic Appearance Flow (DIAF) proposed by Li \textit{et al.} and Liquid Warping GAN (LWGAN) proposed by Liu \textit{et al.} They reported their results on the market-1501~\cite{Zheng2015ScalablePR} dataset and the DeepFashion~\cite{Liu2016DeepFashionPR} dataset. Since our method focuses on high-resolution pose transfer, we only compare with them on the DeepFashion dataset, which has relatively high resolution ($256^2$). Figure~\ref{fig:sota_comparison} shows a qualitative comparison. Our method successfully transfers sharp appearance details to large pose variations and faithfully hallucinates previously occluded parts.  Table~\ref{tab:sota_comparison} shows our better quantitative performance compared with their methods.

\begin{table}[h!]
  \begin{center}
    \caption{Quantitative comparison with previous work.}
    \label{tab:sota_comparison}
\begin{adjustbox}{max width=0.6\linewidth}
    \begin{tabular}{l|c|c|c} 
  \hline
  \multicolumn{3}{c}{DeepFashion}\\
  \textbf{Model} & \textbf{SSIM} & \textbf{MS-SSIM} &\textbf{LPIPS}\\
  \hline
  DSC & 0.776 & 0.792 &0.345\\
  DIAF & 0.778 &0.798 &0.252\\
  LWGAN & 0.781 &0.788 &0.227\\
  Ours & \textbf{0.806} & \textbf{0.814}&\textbf{0.198}\\
  \hline
  Real Data & 1.00 & 1.00 &0.00
\end{tabular}

\end{adjustbox}
  \end{center}
\vspace{-0.3in}
\end{table}

\subsection{Further Applications}
\begin{figure}[t]
    \input{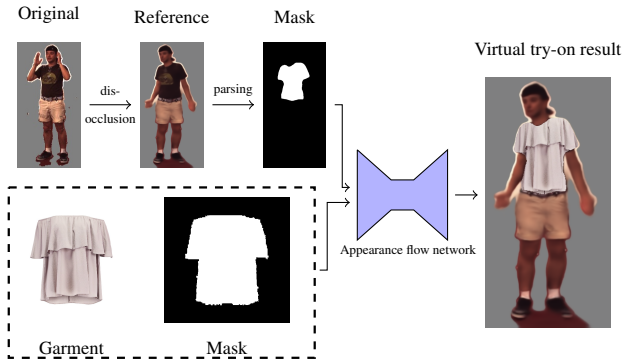}
    \caption{\textbf{Garment Transfer Network.} We first use our Our method to dis-occlude the original image and apply human parsing network~\cite{Liang2018LookIP} to extract the mask of the garment in reference and target poses. The appearance flow network will then take the garment image, reference pose mask and target pose mask as input and outputs the appearance flows, which yield the synthesized garment image through a bilinear sampling layer.}
    \label{fig:garment_transfer}
    \vspace{-0.2in}
\end{figure}

\noindent\textbf{Virtual try-on} Virtual try-on has demonstrated great application potential. This task requires the transfer of any garment with detailed texture. While a recent approach~\cite{VITON} successfully preserves garment details and shapes, there still exists artifacts due to self-occlusions. We can tackle this problem with two steps. First, we transfer the image of the target person (with self-occlusion) into a pre-defined frontal pose (without occlusion). Then, we apply our appearance flow network to transfer the garment to that person. Details are in Figure~\ref{fig:garment_transfer}.\par

\begin{figure}[t]
    \input{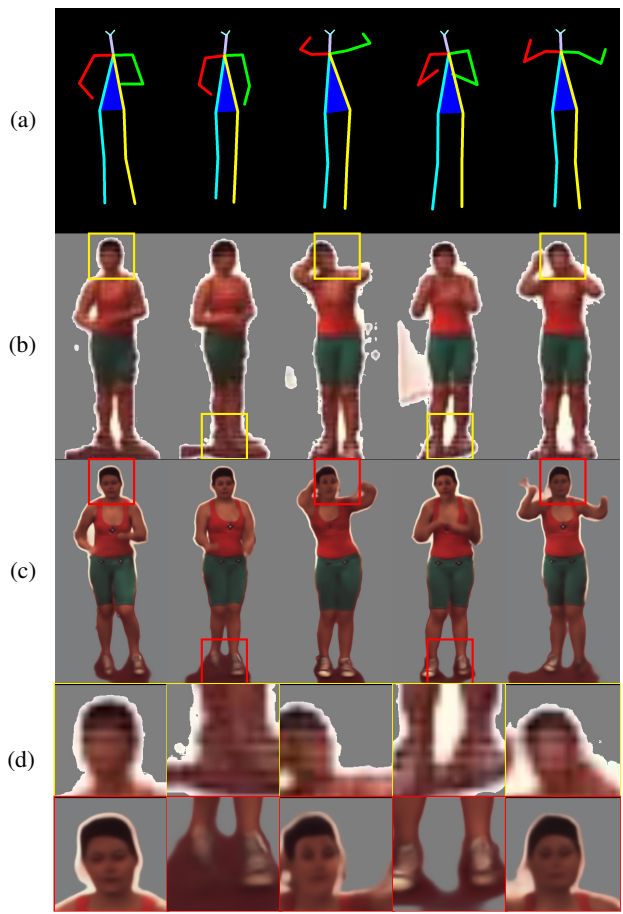}
    \caption{\textbf{Pose-guided human video generation.} five sample frames of (a) input target poses, (b) generated video by~\cite{CaiDeep}, (c) our generated video,  (d) zoom-in views of the two methods.}
    \label{fig:video}
    \vspace{-0.2in}
\end{figure}

\noindent\textbf{Pose-guided human video generation}
Our method can be applied to generate human action {\em videos} in $1024^2$ under a pose guidance. Specifically, given a reference image and a video sequence of target poses, a high-resolution video of the reference person complying the target pose frames is generated. Although we generate the video frame-by-frame and do not consider temporal conherency as done in~\cite{CaiDeep} on human video generation, their video resolution was only $128^2$. Figure~\ref{fig:video} shows a comparison on 5 subsequent frames, where our result shows the facial features as well as other details of the garment and shoes. Our method demonstrates a great potential in high-resolution human video generation.

\section{Conclusion}
In this paper we present a solution for pose-guided high-resolution appearance transfer between images, where the proposed local descriptors (local perceptual loss + local discriminators) and progressive training on autoencoder are shown to be effective in generating plausible and photorealistic images of human at target poses. Self-comparisons have clearly validated the advantages of different components. We have also demonstrated our method's high generalizability in our extensive experiments. We have also shown important applications in high-quality garment transfer and pose-guided human video generation.
{\small
\bibliographystyle{ieee}
\bibliography{egbib}
}

\end{document}